\pdfoutput=1
\documentclass[11pt]{article}
\usepackage[preprint]{acl}
\usepackage{times}
\usepackage{latexsym}
\usepackage[T1]{fontenc}
\usepackage[utf8]{inputenc}
\usepackage{microtype}
\usepackage{graphicx}
\usepackage{booktabs}
\usepackage{multirow}
\usepackage{amsmath,amssymb}
\usepackage{xcolor}

\newcommand{\rcr}{\textsc{RCR}}
\newcommand{\pt}[1]{{\scriptsize(#1)}}

\newcommand{\up}[1]{{\tiny\,#1}}

\title{Drift Says When, Budget Says How Much:\\Rate-Controlled KV Re-selection for Long-Form Multimodal Generation}

\author{Bin Hu \\
  Tsinghua University \And
  Yingbo Xiao \\
  The Chinese University of Hong Kong, Shenzhen}

\hypersetup{pdftitle={Drift Says When, Budget Says How Much: Rate-Controlled KV Re-selection for Long-Form Multimodal Generation},pdfauthor={Bin Hu, Yingbo Xiao}}
\begin{document}
\maketitle

\begin{abstract}
KV cache compression for multimodal LLMs usually selects the visual tokens to keep once, at prefill, and reuses that selection for the whole generation.
We show that this one-shot selection is severely limited on long-form generation: on detailed image description, even an oracle that selects with the attention of the uncompressed answer reaches only 0.44--0.52 ROUGE-L agreement with the uncompressed output at 32 retained prefix tokens per head, because the evidence the model reads moves as the description unfolds.
Re-selecting whenever attention \emph{drifts} out of the kept set is the natural remedy, but we find that a fixed drift threshold fails systematically: drift magnitudes are not comparable across samples, so low-drift samples are almost never refreshed although they benefit the most.
We propose \textbf{Rate-Controlled Refresh} (\rcr), which lets drift decide \emph{when} and \emph{where} to re-select while a per-sample controller regulates \emph{how often}, combined with a low-rank drift detector, candidate-restricted re-selection, and a decision that stays on the GPU.
With one setting for all models and tasks, \rcr{} outperforms the per-step retrieval method InfiniGen by 1.6--3.8 ROUGE-L points on five of six model--task settings with no significant difference on the sixth, while staying within 1.7 points of exact per-step re-selection, with comparable estimated data movement per step and a 4$\times$ smaller GPU-resident detector.
On DetailCaps, drift-chosen timing beats random timing at the same refresh rate and periodic timing even at a higher rate, by 2.0--5.0 points; the advantage over InfiniGen is largest at tight budgets and shrinks as the budget grows.
On pre-registered video description, where drift is homogeneous across samples, \rcr{} is on par with InfiniGen on Qwen2-VL and exceeds it on Qwen3-VL but no longer beats a fixed threshold at the nearest tested rates, consistent with the diagnosis; object-level evaluation shows that \rcr{} achieves higher object F1 against the uncompressed output than InfiniGen on two of three models.
\end{abstract}

\section{Introduction}

The KV cache of a multimodal large language model (MLLM) is dominated by visual tokens: in our long-form benchmarks the prefix holds 640--2,300 tokens per sample, almost all of them image patches.
Prefill-time eviction methods such as SnapKV~\citep{snapkv2024}, PyramidKV~\citep{pyramidkv2024}, Ada-KV~\citep{adakv2025} and the multimodal BACON~\citep{bacon2026} score prefix tokens with the attention of an observation window and keep only the top few.
On visual question answering one-shot selection can be close to lossless: with 64 KV entries per head, an oracle that scores with the true answer's attention matches the uncompressed model on TextVQA and DocVQA.

\begin{figure*}[t]
  \centering
  \includegraphics[width=\textwidth]{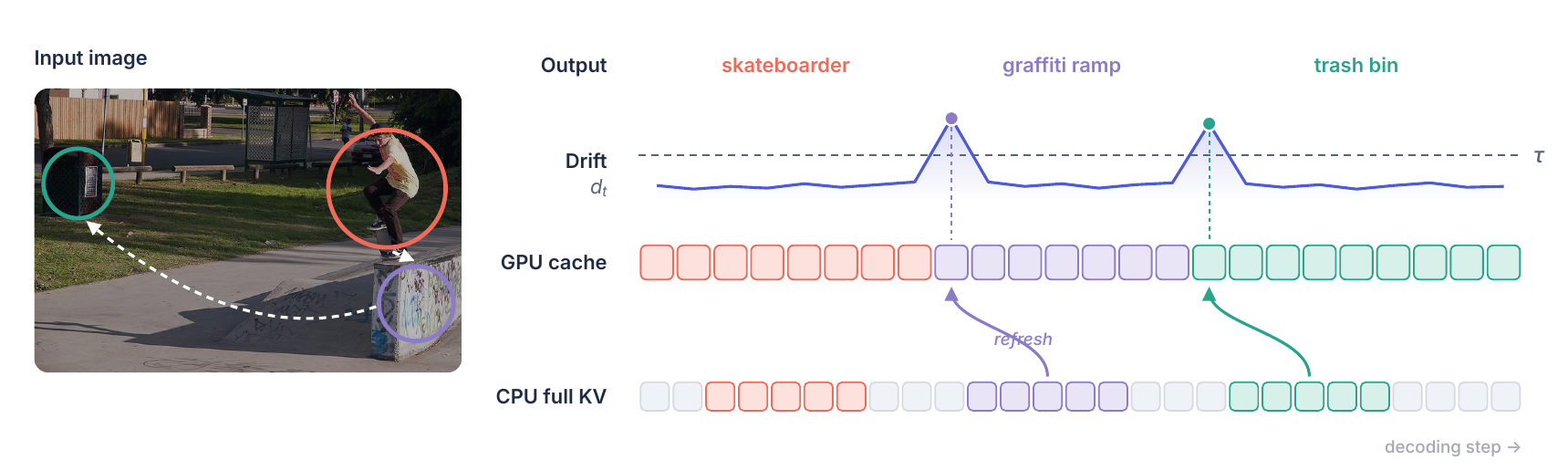}
  \caption{Illustration (not measured). While describing the image, attention moves from the skateboarder to the ramp and the trash bin. Drift $d_t$, the attention mass on evicted prefix tokens, stays low most of the time; when it crosses the threshold, the current query re-selects tokens from the full KV kept on the CPU and overwrites the GPU cache. In \rcr{} the threshold is adapted online per sample so that a fraction $\rho$ of layer-steps refresh.}
  \label{fig:method}
\end{figure*}

Long-form generation is different.
On detailed captioning (DetailCaps;~\citealp{detailcaps2024}) better scoring helps but cannot close the gap: even the answer-attention oracle keeps only 0.44--0.52 ROUGE-L agreement with the uncompressed output (\S\ref{sec:ceiling}).
The reason is that attention \emph{moves}: as the model describes the skateboarder, then the ramp, then the trash bin (Figure~\ref{fig:method}), the evidence it needs leaves the set chosen at prefill.
With 32 slots per head, even an oracle-scored one-shot selection falls well short.

A natural remedy is to keep the full KV off the GPU and re-select whenever the attention mass on evicted tokens, which we call \emph{drift}, exceeds a threshold.
This already beats periodic refresh at equal refresh rates (Appendix~\ref{app:fixed}), yet it trails InfiniGen~\citep{infinigen2024}, which re-selects at every step, by 1.9 points on Qwen2-VL DetailCaps.
A per-sample analysis (\S\ref{sec:diag}) shows where the gap comes from: on high-drift samples the threshold rule already matches exact per-step re-selection, but on low-drift samples it refreshes 1\% of the time while exact re-selection would gain 30 points over BACON.
Drift tells \emph{when} within a sample, but its magnitude is not comparable \emph{across} samples.

We therefore decouple the two decisions.
\rcr{} keeps drift as the signal for when and in which layer to re-select, but adapts the threshold online per sample so that a target fraction $\rho$ of layer-steps refresh.
Three further components make refreshing accurate and cheap: drift is estimated from 8 SVD-rotated key channels, re-selection scores exactly only the detector's top candidates, and the trigger decision never leaves the GPU.
Our contributions are:
\begin{itemize}\setlength\itemsep{1pt}
  \item We quantify the limits of one-shot selection on long-form multimodal generation across three MLLMs (\S\ref{sec:ceiling}).
  \item We identify why fixed drift thresholds fail: the gap to per-step retrieval comes almost entirely from low-drift samples (\S\ref{sec:diag}); on video, where almost no sample is starved, the advantage of rate control over a fixed threshold disappears, consistent with this diagnosis.
  \item We propose \rcr{}, which separates \emph{when} (drift) from \emph{how much} (a per-sample rate budget). One configuration outperforms InfiniGen on five of six model--task settings with no significant difference on the sixth, at comparable CPU traffic; with everything else fixed, drift-chosen timing beats random timing at the same rate and periodic timing even at a higher rate by 2.0--5.0 points (\S\ref{sec:exp}).
  \item We follow a develop-then-confirm protocol: components were chosen on a 100-sample development split and confirmed on held-out data under a pre-registered decision rule (Appendix~\ref{app:protocol}).
\end{itemize}

\section{Related Work}

\paragraph{One-shot KV eviction.}
H2O~\citep{h2o2023} and Scissorhands~\citep{scissorhands2023} keep heavy hitters under a persistence-of-importance assumption; SnapKV~\citep{snapkv2024} votes with an observation window, and PyramidKV~\citep{pyramidkv2024}, Ada-KV~\citep{adakv2025}, HeadKV~\citep{headkv2025} and CAKE~\citep{cake2025} redistribute budgets across layers and heads.
Multimodal variants exploit modality structure~\citep{lookm2024,elasticcache2024,vlcache2025,meda2025,aircache2025,madakv2025,sparsemm2025,mixkv2026}; BACON~\citep{bacon2026} is a strong low-budget selector that we use at prefill.
Look-ahead methods~\citep{laq2025,speckv2026,lookaheadkv2026} estimate future queries before a single eviction.
All of them select once; we show this limits long-form fidelity even with oracle scores.
Visual token reduction prunes or merges image tokens inside the model~\citep{fastv2024,sparsevlm2025,pdrop2025,visionzip2025}, with dynamic or recoverable variants for generation~\citep{dyrate2025,reroute2026}; these change the computation itself and are complementary to KV selection.
For long outputs, several methods keep updating a compressed cache as text is generated~\citep{scope2025,morphkv2025,gkv2025,lazyeviction2025,rkv2025}; \rcr{} instead re-selects among all prefix tokens from a full copy, so visual evidence evicted earlier can return.

\paragraph{Decode-time retrieval.}
Another line keeps the full KV (often offloaded) and retrieves a subset at every step: Quest~\citep{quest2024} bounds page scores with channel-wise key extrema, ArkVale~\citep{arkvale2024} recalls pages by bounding-sphere digests, and InfiniGen~\citep{infinigen2024} speculates attention from a few SVD-skewed key channels; related designs use low-rank keys, approximate nearest neighbours or hashing~\citep{sparq2024,loki2024,retrievalattn2024,magicpig2025,shadowkv2025,specache2025}, share selections across layers~\citep{tidaldecode2025,omnikv2025}, or filter coarse-to-fine~\citep{rocketkv2025}.
Twilight~\citep{twilight2025} replaces fixed top-$k$ budgets with top-$p$ pruning, and ParisKV~\citep{pariskv2026} makes retrieval robust to query drift; both target per-step sparse attention rather than deciding when to re-select.
\rcr{} borrows InfiniGen's skewing for its detector and coarse-to-fine filtering for re-selection, but retrieves only on a budgeted fraction of layer-steps chosen by drift.

\paragraph{Deciding when to recompute.}
ReSA~\citep{resa2025}, Tactic~\citep{tactic2025} and RefreshKV~\citep{recycled2024} recompute at fixed intervals; dynamic variants trigger on query similarity~\citep{recycled2024,retopk2026}, and DSTP~\citep{dstp2026} triggers when the attention distribution over kept visual tokens departs from a prefill anchor.
All use fixed thresholds.
Our trigger signal is the attention mass on evicted tokens, which directly bounds the eviction error~\citep{adakv2025,trustmass2026} and is the online counterpart of QEvict's offline Future Missed Mass~\citep{qevict2026}; our analysis shows that its threshold must adapt per sample.

\section{The Limits of One-Shot Selection}
\label{sec:ceiling}

\paragraph{Setup.}
We use Qwen2-VL-7B-Instruct~\citep{qwen2vl2024}, LLaVA-NeXT-Mistral-7B~\citep{llavanext2024} and Qwen3-VL-8B-Instruct~\citep{qwen3vl2025} with FlashAttention-2.
The budget is 64 KV entries per KV head: $c=32$ selected prefix tokens plus the last $W=32$ observation-window tokens; generated tokens are appended.
Tasks are TextCaps~\citep{textcaps2020} (500 validation images, one-sentence captions, $\le$64 new tokens) and DetailCaps~\citep{detailcaps2024} (300 images, median 157--238 words, $\le$512 new tokens).
All runs use greedy decoding (details in Appendix~\ref{app:impl}). Our metric is \textbf{fidelity}: ROUGE-L F1 between the compressed and the uncompressed output, which isolates the deviation caused by compression; confidence intervals are paired bootstrap 95\%.
Reference-based CIDEr and ROUGE-L separate one-shot from dynamic methods but not dynamic methods from each other (\S\ref{sec:exp}, Appendix~\ref{app:ref}).

\begin{table}[t]
\centering
\small
\setlength{\tabcolsep}{4pt}
\begin{tabular}{lccc}
\toprule
DetailCaps fidelity & Qwen2 & LLaVA & Qwen3 \\
\midrule
SnapKV & .389 & .362 & -- \\
BACON & .388 & .371 & .409 \\
Last-query & .367 & .386 & -- \\
Look-ahead 8 steps & .448 & .412 & -- \\
Prefill oracle (answer attn.) & .516 & .440 & .507 \\
\midrule
Exact re-selection every step & .613 & .478 & .571 \\
\bottomrule
\end{tabular}
\caption{One-shot selection falls short on long descriptions even with oracle scores; re-selecting during decoding is not. --: not run.}
\label{tab:ceiling}
\end{table}

\paragraph{Better scoring is not enough.}
Table~\ref{tab:ceiling} compares one-shot selectors on DetailCaps.
SnapKV, BACON and last-query scoring differ by at most 2.4 points; selecting with the attention of the first 8 answer tokens (decoded with full KV) adds 4--6 points over BACON, and even the prefill oracle, which scores with the uncompressed answer's own attention, stays at .44--.52.
Re-selecting at every decoding step with the current query reaches .48--.61.
On short-answer VQA the picture is the opposite: the prefill oracle loses no accuracy (Qwen2 TextVQA~\citep{textvqa2019} .851 vs.\ .848 uncompressed; similarly on DocVQA~\citep{docvqa2021}), so the gap there is one of scoring.
In long-form generation, we attribute the gap that remains even under oracle scoring to the \emph{staleness} of the selection.
Attention trajectories confirm this: the most-attended prefix position stays the same between adjacent steps only 41\% (Qwen2) to 80\% (LLaVA) of the time, and 37--44\% of newly attended evidence in LLaVA lies far from all previous evidence.

\section{Drift and Why Fixed Thresholds Fail}
\label{sec:diag}

\paragraph{Drift.}
Let $\mathcal{S}_{l,h}$ be the $c$ prefix tokens kept for KV head $h$ of layer $l$, and $a_{t,j}$ the attention of step $t$ on prefix token $j$, averaged over the query heads of the GQA group and normalised over the full prefix plus the rest of the cache.
Drift is the attention mass the model would put on evicted prefix tokens,
\begin{equation}
d_{l,h}(t)=\!\!\sum_{j\notin\mathcal{S}_{l,h}}\!\! a_{t,j},\quad d_l(t)=\frac{1}{H_{kv}}\sum_h d_{l,h}(t).
\end{equation}
A \emph{fixed-threshold} refresh re-selects layer $l$ whenever an estimate $\hat d_l(t)$ exceeds $\tau$, using the current query's exact attention over the full prefix.
With a cheap page-mean estimator (keys averaged over 16-token pages) this rule beats periodic refresh by 3.0--4.2 points at equal refresh rates on Qwen2-VL and LLaVA, but it needs a per-setting $\tau\in[0.02,0.15]$ and trails InfiniGen on Qwen2-VL DetailCaps (.551 vs.\ .570).

\begin{figure}[t]
  \centering
  \includegraphics[width=\columnwidth]{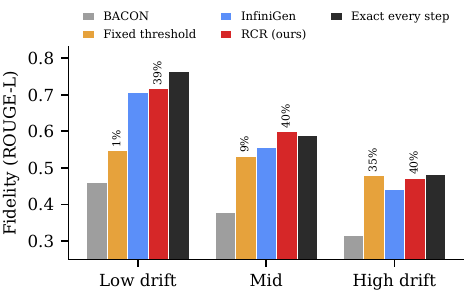}
  \caption{Qwen2-VL DetailCaps split into terciles by each sample's mean drift under the fixed threshold ($\tau=0.03$). Numbers above bars are refresh rates. The fixed threshold matches exact re-selection on high-drift samples but starves low-drift ones; \rcr{} closes this gap.}
  \label{fig:tercile}
\end{figure}

\paragraph{Where the gap comes from.}
Figure~\ref{fig:tercile} splits the 300 Qwen2-VL DetailCaps samples into terciles by their mean drift under $\tau=0.03$.
On \textbf{high-drift} samples the fixed threshold refreshes 35\% of layer-steps and already matches exact per-step re-selection (.479 vs.\ .484), ahead of InfiniGen (.442).
On \textbf{low-drift} samples it refreshes only 1\% of the time, yet these samples gain the most from refreshing (BACON .462, exact .763), and InfiniGen leads by 16 points.
Across samples, the gap to InfiniGen is unrelated to output length (Spearman $-0.01$) but clearly related to mean drift ($-0.41$; Appendix~\ref{app:diag}).
Normalising drift by the total prefix attention does not help: the two quantities have a sample-level rank correlation of 0.98.
In low-drift samples little attention reaches evicted tokens, but that little matters.

\paragraph{Takeaway.}
Within a sample, drift is a good signal of \emph{when} attention has moved; across samples its magnitude does not reflect how valuable a refresh is.
A single threshold therefore cannot allocate refreshes well, which motivates controlling the refresh budget per sample.

\section{Rate-Controlled Refresh}
\label{sec:method}

\paragraph{Cache layout.}
Prefill selects $\mathcal{S}_{l,h}$ with BACON. The full prefix KV ($N_p$ tokens) is kept aside (in CPU memory in a deployment), and the GPU cache is laid out as [$c$ selected $|$ $W$ window $|$ generated]. A refresh overwrites only the first $c$ slots; keys carry RoPE already, so no re-encoding is needed.

\paragraph{Per-sample rate control.}
\rcr{} (Algorithm~1 in Figure~\ref{alg:rcr}) maintains one threshold $\tau$ per sample, shared by all layers and initialised to $\tau_0=0.03$.
After layer $l$ checks step $t$ and refreshes a share $f\in[0,1]$ of its heads, the threshold is updated multiplicatively,
\begin{equation}
\tau \leftarrow \tau\cdot\exp\!\big(\eta\,(f-\rho)\big),\qquad \eta=0.05 ,
\label{eq:rate}
\end{equation}
a constant-gain Robbins--Monro step in $\log\tau$ that is stationary only when the trigger probability equals $\rho$, i.e.\ it tracks the slowly changing $(1-\rho)$-quantile of drift; a persistent trigger excess of 0.1 moves $\tau$ by 15--20\% per decoding step (28--36 layers).
Because $\tau$ is shared across layers, layers that drift more at a given moment still refresh preferentially; because it is reset for every sample, each sample receives roughly a $\rho$ share of refreshes and low-drift samples are no longer starved.
Within a step, layers update $\tau$ in forward order, so later layers see a threshold adjusted by earlier layers' triggers; we did not explore per-layer thresholds. We use $\rho=0.4$ throughout.

\paragraph{Low-rank drift detector.}
At the end of prefill we take, per layer and KV head, the top $R=8$ right singular vectors $P_R\in\mathbb{R}^{128\times R}$ of the prefix key matrix $K\in\mathbb{R}^{N_p\times 128}$ and keep $KP_R$ on the GPU.
Logits of evicted tokens are approximated by $(qP_R)(kP_R)^\top$ while kept tokens use exact logits.
This costs the same as the page-mean estimator ($\approx N_p/16$ dot products per step) but resolves individual tokens and yields a token-level ranking.
The skewing idea follows InfiniGen, which uses 32 channels for per-step selection; \rcr{} uses 8 channels and only for detection and coarse ranking.

\paragraph{Candidate re-selection.}
When a layer triggers, \rcr{} fetches the keys of the detector's top $4c$ tokens, computes their exact attention, keeps the top $c$, and fetches their values.
A refresh thus reads $5c$ vectors instead of $N_p+2c$ for a full-prefix scan.

\paragraph{GPU-side decision.}
A naive implementation copies the drift-vs-threshold comparison to the host in every layer and step to branch on it, which stalls the kernel pipeline.
\rcr{} instead computes the candidate selection in every layer and merges it with the old one under the trigger mask (\texttt{torch.where}); un-triggered heads write back their unchanged entries, so the cache contents are identical to the branching implementation, which we verify token-for-token (Appendix~\ref{app:async}).
Offloading then requires a read conditioned on the mask (\S\ref{sec:disc}).

\begin{figure}[t]
\small
\fbox{\begin{minipage}{0.95\columnwidth}
\textbf{Algorithm 1:} \rcr{} at decoding step $t$, layer $l$\\[-0.6em]
\begin{enumerate}\setlength\itemsep{0pt}\setlength\leftskip{-0.8em}
  \item $\hat a \leftarrow$ detector attention over the prefix, from logits that are exact for $\mathcal{S}_{l,h}$ and $(qP_R)(k_jP_R)^\top$ otherwise
  \item $\hat d_{l,h} \leftarrow \sum_{j\notin\mathcal{S}_{l,h}} \hat a_j$;\quad $m \leftarrow [\tfrac{1}{H_{kv}}\sum_h \hat d_{l,h} > \tau]$
  \item $\tau \leftarrow \tau\exp(\eta(m-\rho))$
  \item $\mathcal{C}_h \leftarrow$ top-$4c$ of $\hat a$;\quad $\mathcal{S}'_{l,h} \leftarrow$ top-$c$ of exact $q k_j^\top$, $j\in\mathcal{C}_h$
  \item $\mathcal{S}_{l,h} \leftarrow m\,?\,\mathcal{S}'_{l,h} : \mathcal{S}_{l,h}$\quad (masked, no host sync)
  \item write K/V of $\mathcal{S}_{l,h}$ into the first $c$ cache slots; run attention
\end{enumerate}
\end{minipage}}
\caption{One \rcr{} step. $\tau$ is shared by all layers and reset to $\tau_0$ for every sample; with layer-level triggering, $m\in\{0,1\}$ and $f=m$ in Eq.~\eqref{eq:rate}.}
\label{alg:rcr}
\end{figure}

\paragraph{Cost.}
Per KV head and step, \rcr{} performs $\approx N_p/16$ detector dot products and reads on average $5c\,r=160\,r$ vectors from the CPU, where $r\approx\rho$ is the refresh rate.
Quest, ArkVale and InfiniGen read $2c=64$ vectors per step, two-stage Quest $5c=160$, and the fixed-threshold rule $r(N_p+2c)$.
The GPU-resident detector of \rcr{} holds $N_p/16$ key equivalents, versus $N_p/4$ for InfiniGen's 32 channels and for Quest's page extrema at page size 8.

\section{Experiments}
\label{sec:exp}

\paragraph{Baselines.}
All dynamic methods start from the same BACON selection and budget, and re-select with the current step's attention only.
\emph{Exact every step} re-selects from exact full-prefix attention at each step; it is the strongest current-query re-selection, not a strict bound (\rcr{} slightly exceeds it on LLaVA DetailCaps).
\emph{Periodic} re-selects at step 1 and every $T$ steps.
\emph{Quest}, \emph{ArkVale} and \emph{two-stage Quest} (page size 8; the latter scores $4c$ candidates exactly) retrieve at every step, as does \emph{InfiniGen} (previous-layer input as query proxy, 32 of 128 skewed channels).
\emph{Fixed threshold} is the page-mean drift rule of \S\ref{sec:diag} with $\tau$ chosen per setting by the rule we registered for Qwen3-VL (the most faithful threshold with a refresh rate of 10--30\%; Appendix~\ref{app:q3}). It also differs from \rcr{} in its detector and full-prefix refresh (separated in Table~\ref{tab:ablation}).

\paragraph{Protocol.}
\rcr{}'s components were selected on a development split of 100 Qwen2-VL DetailCaps images that overlaps the 300-image test set in only two images (Appendix~\ref{app:protocol}).
Three candidates were then evaluated on all six model--task settings under a decision rule registered in advance; we report \rcr{} here and all candidates in Appendix~\ref{app:confirm}.

\begin{table*}[t]
\centering
\small
\setlength{\tabcolsep}{5.6pt}
\begin{tabular}{lcccccc}
\toprule
& \multicolumn{2}{c}{Qwen2-VL-7B} & \multicolumn{2}{c}{LLaVA-NeXT-7B} & \multicolumn{2}{c}{Qwen3-VL-8B} \\
\cmidrule(lr){2-3}\cmidrule(lr){4-5}\cmidrule(lr){6-7}
Method & DetailCaps & TextCaps & DetailCaps & TextCaps & DetailCaps & TextCaps \\
\midrule
BACON (one-shot) & .388 & .626 & .371 & .680 & .409 & .505 \\
Prefill oracle (one-shot) & .516 & .774 & .440 & .810 & .507 & .735 \\
\midrule
Periodic, $T{=}4$ & .521 \pt{25\%} & .738 \pt{28\%} & .448 \pt{25\%} & .751 \pt{29\%} & .535 \pt{25\%} & .687 \pt{27\%} \\
Quest, page 8 & .408 & .644 & .419 & .735 & .440 & .565 \\
ArkVale, page 8 & .347 & .535 & .401 & .697 & .424 & .510 \\
Two-stage Quest, page 8 & .508 & .763 & .468 & .808 & .550 & .684 \\
InfiniGen & .570 & .779 & .451 & .773 & .560 & .733 \\
Fixed threshold & .551 \pt{25\%} & .780 \pt{29\%} & .465 \pt{14\%} & .782 \pt{28\%} & .517 \pt{18\%} & .745 \pt{44\%} \\
\textbf{\rcr{}} ($\rho=0.4$) & \textbf{.598} \pt{40\%} & \textbf{.816} \pt{48\%} & \textbf{.485} \pt{40\%} & \textbf{.811} \pt{48\%} & \textbf{.565} \pt{40\%} & \textbf{.750} \pt{41\%} \\
\quad $\Delta$ vs.\ InfiniGen & +2.8\up{[0.8,4.8]} & +3.7\up{[2.1,5.4]} & +3.4\up{[2.1,4.8]} & +3.8\up{[1.8,5.9]} & +0.5\up{[$-$1.1,2.1]} & +1.6\up{[0.2,3.2]} \\
\midrule
Exact every step & .613 & .826 & .478 & .822 & .571 & .767 \\
\bottomrule
\end{tabular}
\caption{Fidelity (ROUGE-L to the uncompressed output) at a budget of 64 prefix KV entries per head (generated tokens are kept). Parentheses: share of layer-steps that re-select (per-step methods: 100\%). Bold: best method other than exact every step. $\Delta$: paired difference in points with 95\% bootstrap CI.}
\label{tab:main}
\end{table*}

\paragraph{Main results.}
Table~\ref{tab:main} reports fidelity on all six settings.
\textbf{Apart from exact per-step re-selection, \rcr{} is the best method in every setting.}
It improves over BACON by 11.4--21.0 points on DetailCaps and stays within 1.7 points of exact per-step re-selection in every setting (and above it on LLaVA DetailCaps), while refreshing 40--48\% of layer-steps.
\textbf{It outperforms InfiniGen} with confidence intervals above zero on five settings (+1.6 to +3.8), and \textbf{is on par with it on Qwen3-VL DetailCaps} (+0.5, interval including zero).
\textbf{One $\rho$ fits all settings.} The fixed-threshold rule needs a tuned $\tau$ per setting, and on Qwen3-VL its pre-registered setting falls well behind the per-step methods (Appendix~\ref{app:q3}); \rcr{} uses the same configuration everywhere and improves over it by 2.0--4.8 points on DetailCaps.
On TextCaps the realised refresh rate (41--48\%) slightly exceeds $\rho$ because outputs last only 12--29 steps, too few for the threshold to converge.
Page-level retrieval is ineffective at this budget: with only four pages per step, Quest gains at most 6.0 points over BACON and ArkVale falls below BACON on Qwen2-VL.

\begin{figure*}[t]
  \centering
  \includegraphics[width=\textwidth]{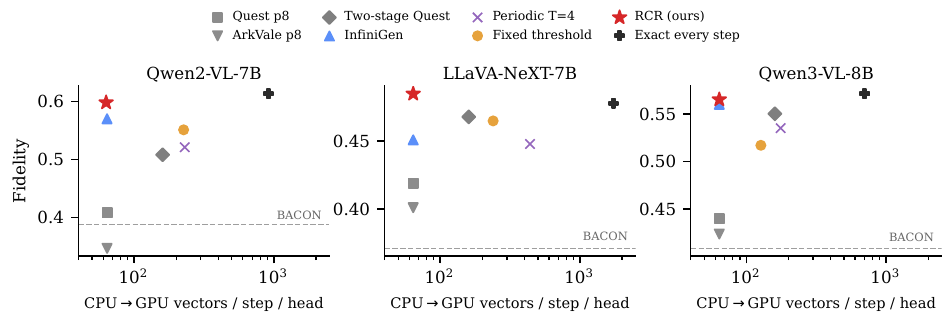}
  \caption{Fidelity vs.\ average number of 128-d vectors read from CPU per step and KV head on DetailCaps (log scale; cost model of \S\ref{sec:method}). At the traffic of a per-step method that fetches only the selected tokens, \rcr{} reaches the highest fidelity.}
  \label{fig:pareto}
\end{figure*}

\paragraph{Cost.}
Figure~\ref{fig:pareto} places each method by its CPU-to-GPU traffic.
\rcr{} reads 63--77 vectors per step and head (160\,$r$), on par with InfiniGen and Quest (64), 39--48\% of two-stage Quest (160), and at most half of the fixed-threshold rule, which scans the full prefix at every trigger.
Its GPU-resident detector is 4$\times$ smaller than InfiniGen's, and it changes the cache on only $\sim$40\% of layer-steps.
The one-off extra cost is an SVD of an $N_p\times128$ matrix per layer and head at the end of prefill.
Our implementation keeps the full prefix on the GPU to simulate offloading, so we report traffic rather than wall-clock latency; moving the trigger decision onto the GPU alone speeds up our simulation by about 3.4$\times$ (Appendix~\ref{app:async}).

\begin{table}[t]
\centering
\small
\setlength{\tabcolsep}{2.5pt}
\resizebox{\columnwidth}{!}{%
\begin{tabular}{lccc}
\toprule
Variant (dev split) & Refresh & Fid. & $\Delta$ vs.\ IG \\
\midrule
Fixed $\tau{=}.03$, page-mean & 15\% & .498 & $-$5.0 \\
\quad + per-head trigger & 15\% & .536 & $-$1.2 \\
Fixed $\tau{=}.02$, page-mean & 26\% & .535 & $-$1.3 \\
Fixed $\tau{=}.03$, low-rank & 28\% & .548 & 0.0 \\
Fixed $\tau{=}.02$, low-rank + cand. & 37\% & .558 & +0.9 \\
Rate $\rho{=}.2$, page-mean & 20\% & .529 & $-$2.0 \\
\textbf{\rcr{}}: rate $\rho{=}.4$ + low-rank + cand. & 40\% & .584 & +3.5 \\
\midrule
InfiniGen / exact every step & 100\% & .549 / .594 & 0 / +4.5 \\
\bottomrule
\end{tabular}}
\caption{Component analysis on the 100-image development split (Qwen2-VL DetailCaps). $\Delta$: paired difference to InfiniGen in points. Full list with intervals in Appendix~\ref{app:protocol}.}
\label{tab:ablation}
\end{table}

\paragraph{Component analysis.}
Table~\ref{tab:ablation} compares \rcr{}'s components on the development split.
At similar refresh rates the low-rank detector improves over page means by 1.3 points, and adding candidate re-selection still gives high fidelity while cutting each refresh from $N_p+2c$ to $5c$ vectors, which is what makes a 40\% refresh rate affordable.
Rate control alone at $\rho=0.2$ is about 1.6 points better than linearly interpolating the fixed-threshold sweep, but its main value is transfer: in the confirmation runs, the two fixed-threshold candidates that refreshed 33--56\% of layer-steps on the development split drifted to 21--100\% on other models and tasks, whereas \rcr{} stayed at 40--48\% with similar fidelity (Appendix~\ref{app:confirm}).
Per-head triggering helps at low refresh rates but did not transfer as predictably.

\begin{table}[t]
\centering
\small
\setlength{\tabcolsep}{2.4pt}
\begin{tabular}{lccc}
\toprule
Trigger (same detector, cand.) & Refresh & Qwen2 & LLaVA \\
\midrule
Random & 40\% & .548\up{$-$5.0} & .462\up{$-$2.3} \\
Periodic, $T{=}3$ & 33\% & .552\up{$-$4.6} & -- \\
Periodic, $T{=}2$ & 50\% & .578\up{$-$2.0} & -- \\
Drift, rate-controlled (\rcr{}) & 40\% & \textbf{.598} & \textbf{.485} \\
\bottomrule
\end{tabular}
\caption{Does drift decide \emph{when}? DetailCaps fidelity when only the trigger changes; small numbers: paired difference to \rcr{} (all 95\% CIs exclude zero; Appendix~\ref{app:sens}).}
\label{tab:when}
\end{table}

\paragraph{Does drift decide \emph{when}?}
Table~\ref{tab:when} keeps \rcr{}'s detector, candidate re-selection and merge and replaces only the trigger.
At the same 40\% rate, random timing loses 5.0 points on Qwen2-VL and 2.3 on LLaVA-NeXT, and on Qwen2-VL falls below InfiniGen (.548 vs.\ .570); periodic timing loses 2.0 points even when it refreshes more often (50\%).
The gains thus come from refreshing at the right moments, not merely often.

\begin{figure}[t]
  \centering
  \includegraphics[width=\columnwidth]{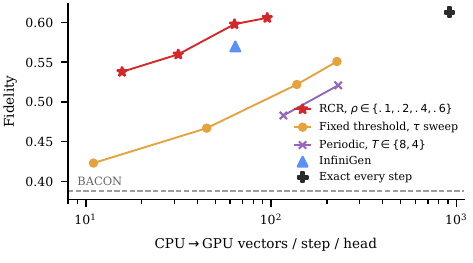}
  \caption{Post-hoc sweeps on Qwen2-VL DetailCaps (300 test images, budget 64). \rcr{} points vary only $\rho$ (the main configuration uses $\rho{=}0.4$, fixed beforehand); fixed-threshold and periodic points vary $\tau$ and $T$ and scan the full prefix at each refresh.}
  \label{fig:sens}
\end{figure}

\paragraph{Sensitivity to $\rho$ and to the budget.}
After the confirmation we swept $\rho$ on the Qwen2-VL DetailCaps test set (Figure~\ref{fig:sens}; full numbers in Appendix~\ref{app:sens}).
Fidelity rises monotonically with $\rho$ (.538, .560, .598, .606 for $\rho=0.1, 0.2, 0.4, 0.6$) and approaches exact per-step re-selection (.613), so $\rho$ acts as a smooth quality--cost knob rather than a fragile hyperparameter.
At $\rho=0.2$ \rcr{} is on par with InfiniGen ($-1.0$ [$-2.9$, $+1.0$]) while reading half as many vectors (32 vs.\ 64 per step and head); at $\rho=0.1$ it trails InfiniGen by 3.2 points [$-5.3$, $-1.0$] but still exceeds the fixed threshold at $\tau=0.03$ (.522) and periodic refresh with $T=4$ (.521), which read 8.6 and 14 times more data.
Larger budgets narrow all gaps: at 128 entries per head ($c=96$), \rcr{} still leads InfiniGen by 2.7 points [+1.3, +4.1] (.785 vs.\ .758; exact .788), but at 256 ($c=224$) the lead shrinks to 1.2 [$-0.1$, +2.4] (.861 vs.\ .849; exact .874, BACON .785), so \rcr{} matters most when the budget is tight.

\subsection{Analysis}

\paragraph{Does rate control fix the diagnosed failure?}
Figure~\ref{fig:tercile} shows \rcr{} on the same terciles.
On low-drift samples it raises fidelity from .548 to .718, above InfiniGen (.707), while on high-drift samples it stays at the level of the fixed threshold (.473 vs.\ .479) and of exact re-selection (.484).
The improvement comes from where the diagnosis predicted.
On LLaVA-NeXT, whose fixed-threshold refresh rate is more uniform across terciles (10--18\%), \rcr{} improves the low and middle terciles (.455$\to$.493, .472$\to$.495) and leaves the high one unchanged (.469 vs.\ .466).

\begin{figure}[t]
  \centering
  \includegraphics[width=\columnwidth]{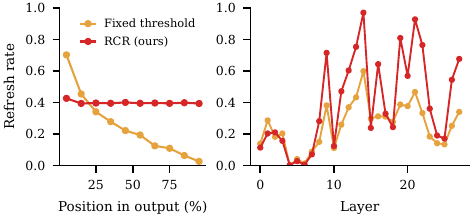}
  \caption{Where refreshes go (61 Qwen2-VL DetailCaps samples, layer-step traces). Left: refresh rate by position in the output. Right: refresh rate per layer. The fixed threshold ($\tau{=}0.03$) spends its refreshes at the start of generation; \rcr{} spreads them evenly over time while keeping the same preference over layers.}
  \label{fig:alloc}
\end{figure}

\begin{figure}[t]
  \centering
  \includegraphics[width=\columnwidth]{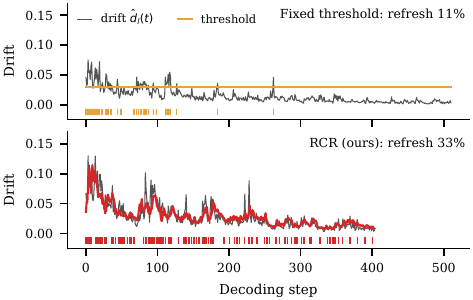}
  \caption{Drift, threshold and refreshes (ticks) for layer 11 on one Qwen2-VL DetailCaps image. \rcr{}'s threshold is replayed exactly from the traced decisions. The fixed threshold stops refreshing once drift decays; \rcr{} keeps refreshing at a steady rate. Fidelity: fixed .36, \rcr{} .55.}
  \label{fig:traj}
\end{figure}

\paragraph{Where does \rcr{} spend its refreshes?}
We traced every layer-step decision on the first 61 Qwen2-VL DetailCaps test images (Figure~\ref{fig:alloc}); tracing leaves all outputs unchanged, and these images drift more than average (fixed-threshold refresh 29\% vs.\ 15\% on the full test set).
Two observations refine the diagnosis.
First, drift is not comparable across \emph{time} either: as generated tokens accumulate they absorb attention, so drift decays and the fixed threshold refreshes 70\% of layer-steps in the first tenth of the output but only 3\% in the last tenth.
Because \rcr{}'s threshold tracks a running quantile, it follows this decay and refreshes 39--43\% of layer-steps in every decile.
Figure~\ref{fig:traj} shows this on one image and layer: after step 130 drift stays below $\tau=0.03$ on 99\% of steps and the fixed rule refreshes only twice more in 380 steps, whereas \rcr{}'s threshold descends with the drift and keeps refreshing (69 times) until the end of the description.
Second, rate control redistributes refreshes across samples and steps but not across layers: the per-layer refresh profiles of the two methods have a Spearman correlation of 0.93, with layers 4--7 almost never refreshing and layers 14 and 21 refreshing on most steps under \rcr{}.
Drift thus keeps deciding \emph{where} to re-select, while the controller decides \emph{how much}.

\paragraph{Why are gains smaller on Qwen3-VL?}
Qwen3-VL drifts less and has shorter prefixes (639 vs.\ 852 tokens), so InfiniGen loses less: exact re-selection is only 1.1 points above it (4.3 on Qwen2-VL), and \rcr{} recovers about half of this.

\paragraph{Video: a test of the diagnosis.}
On 300 pre-registered ActivityNet videos of VideoDetailCaption~\citep{activitynet2015,lmmseval2024} (32 frames, $\sim$2,300 prefix tokens; Appendix~\ref{app:video}), per-sample drift is far more homogeneous than on DetailCaps images (coefficient of variation 0.14--0.18 vs.\ 0.71--0.78 for the Qwen models), and the fixed threshold starves at most 1\% of videos (vs.\ 21--45\% of images for the Qwen models).
\rcr{}, unchanged, is on par with InfiniGen on Qwen2-VL (+0.9 [$-$0.6, +2.4]) and exceeds it on Qwen3-VL (+2.0 [+1.1, +2.8]), but at the nearest tested rates it is no better than the fixed threshold ($-$1.0 to +0.5, all intervals include zero), consistent with the diagnosis.

\paragraph{Reference metrics and object-level errors.}
Against human references, dynamic methods clearly beat BACON; CIDEr and ROUGE-L do not separate them from each other or from the uncompressed model (Appendix~\ref{app:ref}), and CAPTURE, the object-level DetailCaps metric~\citep{detailcaps2024}, separates them only by small margins (.631--.635 on Qwen2-VL; uncompressed .636).
Parsing outputs into objects with CAPTURE's scene-graph parser and comparing them with the uncompressed output (Appendix~\ref{app:capture}), \rcr{} exceeds InfiniGen in object F1 by 1.3 [0.2, 2.3] points on Qwen2-VL and 2.5 [1.4, 3.5] on LLaVA-NeXT, is on par on Qwen3-VL (+0.7 [$-$0.1, 1.5]), and exceeds the fixed threshold by 1.3--2.2.
On LLaVA-NeXT the gain is recall: \rcr{} omits fewer of the uncompressed model's objects (+4.8 [3.3, 6.2]) at similar precision ($-$0.7 [$-$2.1, +0.7]), although its precision against the human references is 2.2 points below InfiniGen's.
Appendix~\ref{app:qual} shows an omission by InfiniGen and a counter-example.

\section{Discussion}
\label{sec:disc}

\paragraph{Why a rate rather than a better threshold?}
Drift is an uncalibrated score: its scale depends on the image, the model, and even the position in the output (Figure~\ref{fig:alloc}).
Any fixed cut-off on such a score yields a refresh rate that varies severalfold across settings (Table~\ref{tab:main}, 14--44\%; Appendix~\ref{app:confirm}, 21--100\%).
Tracking a quantile instead turns the score into a ranking, which is all the trigger needs, and turns the hard-to-set threshold into an interpretable budget.
Other online triggers~\citep{recycled2024,dstp2026} also use fixed thresholds today.

\paragraph{Choosing $\rho$.}
Average traffic is $5c\rho$ vectors per layer, KV head and step, so $\rho$ follows from a bandwidth budget: $\rho=0.4$ matches InfiniGen's 64 vectors, and $\rho=0.2$ halves them at a cost of 3.8 points on Qwen2-VL DetailCaps (Figure~\ref{fig:sens}); a fixed threshold's traffic depends on each model's and task's drift statistics.

\paragraph{Deployment.}
Our simulation keeps the full prefix on the GPU. Offloaded, \rcr{} must read candidate keys only for triggered heads without involving the host, e.g.\ with a kernel that reads pinned host memory through unified virtual addressing under the trigger mask; fetching candidates unconditionally would cost $5c=160$ vectors per step, as two-stage Quest. Unlike InfiniGen, which speculates with the previous layer's input, \rcr{} fetches on the critical path of a refreshing layer; combining the two in a real offloading system is future work.

\section{Conclusion}
One-shot KV selection limits long-form multimodal generation because the model's attention moves over the image; even an oracle keeps only .44--.52 fidelity.
Re-selecting on drift is the natural remedy, but drift magnitudes are not comparable across samples, so a fixed threshold starves exactly the samples that need refreshing.
\rcr{} lets drift decide when and where to re-select and a per-sample budget decide how often.
One configuration outperforms InfiniGen in five of six image settings at comparable traffic, beats random timing at the same refresh rate and periodic timing even when the latter refreshes more often, and exposes a knob that maps to bandwidth; its advantage shrinks with larger budgets and, over a fixed threshold, vanishes on video, where drift is homogeneous.
\section*{Limitations}
\textbf{Scope.} We evaluate two image captioning tasks on three 7--8B MLLMs and video description on two, at a prefix budget of 64 entries per head (generated tokens are kept), and budgets of 128 and 256 only on Qwen2-VL DetailCaps, where the advantage over InfiniGen is no longer significant at 256; larger models, multi-image generation, long document answers, multimodal reasoning chains and short-answer VQA are not covered. On video, \rcr{} is not significantly better than InfiniGen on Qwen2-VL and not better than a fixed threshold at the nearest tested rates on either model; on LLaVA-NeXT images, whose drift is similarly homogeneous, its gain over a fixed threshold at a similar rate is borderline (+1.1 [$-$0.0, +2.3]). The $\rho$ and budget sweeps were run after the confirmation and on the test set; they characterise the method but were not used to choose it.
\textbf{Evaluation.} Fidelity measures deviation from the uncompressed output, not correctness. Reference-based metrics separate dynamic methods weakly at best (CAPTURE differences are below one point); our object-level analysis relies on an automatic scene-graph parser, and we run no human evaluation.
\textbf{Baselines.} All baselines are our re-implementations in a shared simulation; we do not re-implement DSTP~\citep{dstp2026}, whose trigger compares the attention over kept visual tokens with a prefill anchor, so our fixed-threshold baseline covers fixed thresholds on drift but not on DSTP's signal.
\textbf{Gains vary by model.} \rcr{} outperforms InfiniGen by 2.8--3.8 points on Qwen2-VL and LLaVA-NeXT but only by 1.6 on Qwen3-VL TextCaps and not significantly on Qwen3-VL DetailCaps, where drift is uniformly small; the fixed-threshold version is clearly weaker there (Appendix~\ref{app:q3}).
\textbf{Selection on a development split.} Components were chosen among 14 variants on 100 development images, and confidence intervals are not corrected for multiple comparisons; held-out confirmation mitigates but cannot fully remove selection bias, and the margin over InfiniGen shrank from 3.5 on the development split to 2.8 on the Qwen2-VL test set.
\textbf{Latency.} Costs are traffic estimates from a simulation that keeps the full prefix on the GPU; we do not measure end-to-end latency of an offloading system, and \rcr{}'s fetch cannot be prefetched with the previous layer's input as in InfiniGen without modification (\S\ref{sec:disc}).
\textbf{Short outputs.} On 12--29-step outputs the controller overshoots $\rho$ (41--48\%).

\bibliography{refs}

\appendix

\section{Development and Confirmation Protocol}
\label{app:protocol}

\paragraph{Development split.} We compared 14 variants on 100 Qwen2-VL DetailCaps images sampled evenly over the 4,870-image set (indices from \texttt{linspace} with $n=100$; only indices 0 and 4869 also appear in the 300-image test set). Reference methods were re-run on the same images. Table~\ref{tab:dev} lists every variant.

\begin{table*}[t]
\centering
\small
\resizebox{\textwidth}{!}{%
\begin{tabular}{llccc}
\toprule
Tag & Configuration & Refresh & Fidelity & $\Delta$ vs.\ InfiniGen \\
\midrule
\texttt{b64\_bacon} & BACON & -- & .367 & -- \\
\texttt{grow3\_px} & threshold $=3\times$ residual drift after last refresh, $\tau_{\min}{=}.005$ & 0.5\% & .396 & -- \\
\texttt{grow2h\_px} & threshold $=2\times$ residual drift, per head & 0.6\% & .430 & -- \\
\texttt{async\_px003} & fixed $\tau{=}.03$, page-mean (fixed-threshold version) & 15.4\% & .498 & $-$5.0 [$-$8.4, $-$1.8] \\
\texttt{head\_px003} & + per-head trigger & 15.3\% & .536 & $-$1.2 [$-$4.5, +1.9] \\
\texttt{rate20\_px} & rate control $\rho{=}.2$, page-mean & 19.7\% & .529 & $-$2.0 [$-$5.8, +1.6] \\
\texttt{floor8\_px003} & $\tau{=}.03$ + forced refresh after 8 steps & 24.9\% & .530 & -- \\
\texttt{period14} & periodic $T{=}4$ & 25.1\% & .509 & -- \\
\texttt{px002} & fixed $\tau{=}.02$, page-mean & 26.4\% & .535 & $-$1.3 [$-$4.7, +1.9] \\
\texttt{svd8\_003} & $\tau{=}.03$, 8-channel low-rank detector & 27.7\% & .548 & $-$0.0 [$-$3.5, +3.2] \\
\texttt{svd8c4h\_002} & $\tau{=}.02$, low-rank + $4c$ candidates + per head & 32.5\% & .572 & +2.4 [$-$0.5, +5.3] \\
\texttt{svd8c4\_002} & $\tau{=}.02$, low-rank + $4c$ candidates & 36.9\% & .558 & +0.9 [$-$2.3, +4.2] \\
\texttt{rate40\_c4} & \rcr{}: $\rho{=}.4$, low-rank + $4c$ candidates & 39.7\% & .584 & +3.5 [+0.5, +6.7] \\
\texttt{floor4\_px002} & $\tau{=}.02$ + forced refresh after 4 steps & 42.7\% & .572 & +2.4 [$-$0.9, +5.7] \\
\texttt{floor8\_c4\_002} & $\tau{=}.02$, low-rank + candidates + forced after 8 & 43.3\% & .574 & +2.5 [$-$0.1, +5.3] \\
\texttt{svd8c4\_001} & $\tau{=}.01$, low-rank + $4c$ candidates & 55.9\% & .597 & +4.8 [+1.6, +8.1] \\
\texttt{questex8} & two-stage Quest, page 8 & 100\% & .530 & $-$1.9 [$-$4.7, +1.0] \\
\texttt{infinigen} & InfiniGen & 100\% & .549 & 0 \\
\texttt{reveryd0} & exact every step & 100\% & .594 & +4.5 [+1.5, +7.5] \\
\bottomrule
\end{tabular}}
\caption{All development-split results (Qwen2-VL DetailCaps, 100 images), sorted by refresh rate, except that the per-head variant directly follows its base configuration. ``--'': interval not computed.}
\label{tab:dev}
\end{table*}

\paragraph{Selection.} We advanced the most faithful fixed-threshold combination (\texttt{svd8c4\_001}), the cheapest combination of similar fidelity (\texttt{svd8c4h\_002}), and the only combination without a per-setting trigger threshold (\texttt{rate40\_c4}, \rcr{}). Forced refresh (\texttt{floor*}) performed similarly to rate control but adds a hyperparameter. Thresholds relative to the residual drift after the last refresh (\texttt{grow*}) almost never fired.

\paragraph{Pre-registered decision rule.} Before running the confirmation, we fixed the three candidates, the six settings, the comparison to InfiniGen on identical samples, and the rule: a candidate becomes the main method if it is not significantly worse than InfiniGen (upper CI bound $\ge 0$; this is not a non-inferiority test) in at least five of six settings; all results are reported regardless. \rcr{} satisfies the rule in all six settings.

\section{Confirmation Results for All Candidates}
\label{app:confirm}

\begin{table*}[t]
\centering
\small
\begin{tabular}{lccc}
\toprule
Setting & \rcr{} ($\rho{=}.4$) & $\tau{=}.01$ + low-rank + cand. & $\tau{=}.02$ + low-rank + cand. + per head \\
\midrule
Qwen2 DetailCaps & +2.8 [+0.8, +4.8] \pt{40\%} & +2.8 [+0.9, +4.6] \pt{54\%} & +1.9 [$-$0.1, +3.8] \pt{31\%} \\
Qwen2 TextCaps & +3.7 [+2.1, +5.4] \pt{48\%} & +5.1 [+3.3, +6.8] \pt{100\%} & +4.8 [+3.0, +6.6] \pt{92\%} \\
LLaVA DetailCaps & +3.4 [+2.1, +4.8] \pt{40\%} & +3.5 [+2.1, +4.8] \pt{96\%} & +3.4 [+2.1, +4.8] \pt{66\%} \\
LLaVA TextCaps & +3.8 [+1.8, +5.9] \pt{48\%} & +4.4 [+2.4, +6.4] \pt{100\%} & +4.6 [+2.6, +6.7] \pt{92\%} \\
Qwen3 DetailCaps & +0.5 [$-$1.1, +2.1] \pt{40\%} & +0.9 [$-$0.7, +2.6] \pt{40\%} & +1.1 [$-$0.5, +2.9] \pt{21\%} \\
Qwen3 TextCaps & +1.6 [+0.2, +3.2] \pt{41\%} & +3.5 [+2.1, +4.9] \pt{99\%} & +3.6 [+2.1, +5.0] \pt{69\%} \\
\bottomrule
\end{tabular}
\caption{Paired fidelity difference to InfiniGen (points, 95\% CI) and refresh rate for the three confirmed candidates.}
\label{tab:confirm}
\end{table*}

Table~\ref{tab:confirm} lists all three candidates. The two fixed-threshold candidates are as accurate as \rcr{} or slightly more on TextCaps, but at 69--100\% refresh, i.e.\ nearly per-step cost; their refresh rates range from 21\% to 100\% across settings, whereas \rcr{} stays at 40--48\%. Absolute fidelities on the evaluated subsets, together with InfiniGen, two-stage Quest, exact re-selection and BACON on the same samples, are in the released logs.

\section{Pre-registered Qwen3-VL Matrix for the Fixed-Threshold Version}
\label{app:q3}

Before exploring any of \rcr{}'s components, we registered a full Qwen3-VL-8B evaluation of the fixed-threshold version, including the rule for its headline setting (the most faithful threshold with refresh rate in 10--30\%, else the one closest to 20\%). Table~\ref{tab:q3} reports all results. By the rule, the headline settings are $\tau=0.02$ on DetailCaps and $\tau=0.05$ on TextCaps.
On DetailCaps the fixed threshold trails InfiniGen by 4.3 points, two-stage Quest by 3.3 and periodic $T{=}4$ by 1.8; interpolated to a 25\% refresh rate it reaches about +12.3 over BACON, on par with periodic refresh at the same rate (+12.5), i.e.\ the advantage over periodic refresh seen on Qwen2-VL and LLaVA does not appear.
On TextCaps it is 1.2 points above InfiniGen. Qwen3-VL drifts less overall ($\tau{=}0.03$ refreshes only 11\% on DetailCaps), the low-drift regime diagnosed in \S\ref{sec:diag}.
During the run, the $\tau{=}0.03$ job was restarted with the GPU-side decision after 21 samples; its first 21 outputs were identical to the original implementation (Appendix~\ref{app:async}).

\begin{table}[h]
\centering
\small
\setlength{\tabcolsep}{3pt}
\begin{tabular}{lcc}
\toprule
Configuration & DetailCaps & TextCaps \\
\midrule
BACON & .409 & .505 \\
Prefill oracle & .507 & .735 \\
Periodic $T{=}8$ & .507 \pt{13\%} & .633 \pt{14\%} \\
Periodic $T{=}4$ & .535 \pt{25\%} & .687 \pt{27\%} \\
Quest / ArkVale, page 8 & .440 / .424 & .565 / .510 \\
Two-stage Quest, page 8 & .550 & .684 \\
InfiniGen & .560 & .733 \\
Fixed $\tau{=}.05$ & .474 \pt{5\%} & \textbf{.745} \pt{44\%} \\
Fixed $\tau{=}.03$ & .492 \pt{11\%} & .759 \pt{64\%} \\
Fixed $\tau{=}.02$ & \textbf{.517} \pt{18\%} & .765 \pt{79\%} \\
Fixed $\tau{=}.01$ & .552 \pt{34\%} & .766 \pt{94\%} \\
Exact every step & .571 & .767 \\
\bottomrule
\end{tabular}
\caption{Pre-registered Qwen3-VL-8B matrix (all runs complete). Bold: headline settings chosen by the registered rule.}
\label{tab:q3}
\end{table}

\section{Fixed-Threshold Version: Additional Results}
\label{app:fixed}

\paragraph{Versus periodic refresh at equal rates.} On Qwen2-VL and LLaVA the fixed threshold outperforms periodic refresh at matched refresh rates by +3.0 (Qwen2 DetailCaps, 24.7\% vs.\ 25.1\%), +4.2 (Qwen2 TextCaps, 28.9\% vs.\ 28.1\%), +3.6 (LLaVA DetailCaps, 13.7\% vs.\ 12.7\%) and +3.1 (LLaVA TextCaps, 28.2\% vs.\ 28.6\%). Its thresholds were 0.02, 0.12, 0.08 and 0.15 (chosen by the rule of Appendix~\ref{app:q3}), with gains over BACON of +16.3 [14.4, 18.2], +15.4 [12.9, 17.9], +9.5 [8.2, 10.8] and +10.2 [8.1, 12.5].

\paragraph{Page-mean estimator.} On 20--40 calibration samples per setting, page-mean drift has a step-level Spearman correlation of 0.90--0.96 with exact drift, and recalls 0.54--0.91 of the high-drift steps flagged by the exact detector at trigger rates of 3--50\%.

\paragraph{EMA.} Re-selecting from an exponential moving average of past attention (decay 0.8) instead of the current step hurts exact per-step re-selection (e.g.\ +19.4 vs.\ +22.5 over BACON on Qwen2 DetailCaps) and drift-triggered refresh (e.g.\ +6.8 vs.\ +7.9), and changes periodic refresh by at most $\pm$0.8. All results in the paper use the current step only.

\section{Diagnosis Details}
\label{app:diag}
All numbers in this appendix use $\tau=0.03$, as in Figure~\ref{fig:tercile}. On Qwen2-VL DetailCaps InfiniGen beats the fixed threshold (by more than 2 points) on 46\% of samples and loses on 41\%; on TextCaps, where the fixed threshold is ahead by 4.7 points on average, the gap is unrelated to drift (Spearman +0.01). Against two-stage Quest the DetailCaps gap is again concentrated in the low-drift tercile (+4.1 for two-stage Quest) and reversed in the high-drift one ($-$6.2). Exact re-selection beats the fixed threshold by 9.1 points on average; the per-sample gain correlates with refresh rate ($-0.39$) but not with output length ($-0.02$). Relative drift by tercile is 0.025 / 0.049 / 0.094 and total prefix attention 0.265 / 0.295 / 0.330. In the layer-step traces of Figure~\ref{fig:alloc}, each refresh replaces about a third of the 32 kept tokens (35\% for both methods), and per-sample refresh rates span 13--44\% (10th--90th percentile) under the fixed threshold and 39.8--40.1\% under \rcr{}.

\section{GPU-Side Decision: Equivalence and Speed}
\label{app:async}
The GPU-side decision computes the candidate selection in every layer and merges it with the old selection under the trigger mask; un-triggered heads rewrite their own entries, so the cache equals that of the branching implementation. On Qwen3-VL DetailCaps ($\tau=0.03$), the first 21 samples have identical outputs, per-layer refresh counts, scan counts and mean drift under both implementations. On Qwen2-VL, the GPU-side fixed threshold and exact every-step re-selection (implemented as \texttt{drift:-1}) reproduce the original outputs on the two samples shared with the original runs. With 14 evaluation processes on a 4-core host, removing per-layer host synchronisation reduced Qwen3-VL evaluation time from about 137 to 40 seconds per sample; this reflects simulation efficiency, not deployment latency.

\section{Sensitivity Runs}
\label{app:sens}
Table~\ref{tab:sens} lists the post-hoc sweeps of Figure~\ref{fig:sens} with intervals and reference-based ROUGE-L (against the DetailCaps references). Reference ROUGE-L is noisy at this scale and not monotone in $\rho$ (.277 to .285), while fidelity is.

\begin{table}[h]
\centering
\small
\setlength{\tabcolsep}{3pt}
\resizebox{\columnwidth}{!}{%
\begin{tabular}{lcccc}
\toprule
Qwen2-VL DetailCaps & Refresh & Fidelity & $\Delta$ vs.\ IG & Ref.\ R-L \\
\midrule
\multicolumn{5}{l}{\emph{Budget 64}} \\
InfiniGen & 100\% & .570 & 0 & .281 \\
\rcr{} $\rho{=}0.1$ & 9.8\% & .538 & $-$3.2 [$-$5.3, $-$1.0] & .277 \\
\rcr{} $\rho{=}0.2$ & 19.7\% & .560 & $-$1.0 [$-$2.9, +1.0] & .279 \\
\rcr{} $\rho{=}0.4$ & 39.6\% & .598 & +2.8 [+0.8, +4.8] & .285 \\
\rcr{} $\rho{=}0.6$ & 59.5\% & .606 & +3.6 [+1.6, +5.6] & .279 \\
Exact every step & 100\% & .613 & +4.3 [+2.4, +6.1] & .282 \\
\midrule
\multicolumn{5}{l}{\emph{Budget 128}} \\
BACON & -- & .667 & $-$9.1 [$-$10.8, $-$7.4] & .271 \\
InfiniGen & 100\% & .758 & 0 & .284 \\
\rcr{} $\rho{=}0.4$ & 39.7\% & .785 & +2.7 [+1.3, +4.1] & .285 \\
Exact every step & 100\% & .788 & +3.0 [+1.5, +4.6] & .287 \\
\midrule
\multicolumn{5}{l}{\emph{Budget 256}} \\
BACON & -- & .785 & $-$6.4 [$-$8.0, $-$5.0] & -- \\
InfiniGen & 100\% & .849 & 0 & -- \\
\rcr{} $\rho{=}0.4$ & 39.5\% & .861 & +1.2 [$-$0.1, +2.4] & -- \\
Exact every step & 100\% & .874 & +2.5 [+1.2, +3.8] & -- \\
\midrule
\multicolumn{5}{l}{\emph{Budget 64, same detector and candidates, other triggers ($\Delta$ vs.\ RCR)}} \\
Random (Qwen2) & 40.2\% & .548 & $-$5.0 [$-$7.2, $-$2.6] & -- \\
Periodic $T{=}3$ (Qwen2) & 33.5\% & .552 & $-$4.6 [$-$6.5, $-$2.6] & -- \\
Periodic $T{=}2$ (Qwen2) & 50.1\% & .578 & $-$2.0 [$-$3.8, $-$0.2] & -- \\
Random (LLaVA) & 40.3\% & .462 & $-$2.3 [$-$3.7, $-$0.9] & -- \\
\bottomrule
\end{tabular}}
\caption{Post-hoc $\rho$, budget and trigger sweeps on DetailCaps (300 test images; Qwen2-VL unless noted). $\Delta$ is against InfiniGen except in the last block. Reference ROUGE-L was computed only for the first two blocks.}
\label{tab:sens}
\end{table}

\section{Reference-Based Metrics}
\label{app:ref}
Table~\ref{tab:ref} reports corpus CIDEr and ROUGE-L against the dataset references for the main methods on all six settings (same samples as Table~\ref{tab:main}). On DetailCaps, CIDEr is too small to rank methods (e.g.\ .004--.012 on Qwen3-VL).

\begin{table*}[t]
\centering
\small
\setlength{\tabcolsep}{4pt}
\begin{tabular}{lcccccc}
\toprule
CIDEr / ROUGE-L & \multicolumn{2}{c}{Qwen2-VL-7B} & \multicolumn{2}{c}{LLaVA-NeXT-7B} & \multicolumn{2}{c}{Qwen3-VL-8B} \\
\cmidrule(lr){2-3}\cmidrule(lr){4-5}\cmidrule(lr){6-7}
Method & DetailCaps & TextCaps & DetailCaps & TextCaps & DetailCaps & TextCaps \\
\midrule
Uncompressed & .067 / .287 & 1.524 / .550 & .099 / .285 & .777 / .445 & .009 / .261 & .376 / .384 \\
BACON & .044 / .261 & 1.209 / .506 & .073 / .262 & .578 / .416 & .004 / .249 & .318 / .348 \\
Two-stage Quest & .057 / .280 & 1.462 / .545 & .091 / .279 & .741 / .444 & .005 / .256 & .390 / .382 \\
InfiniGen & .061 / .281 & 1.550 / .554 & .099 / .279 & .715 / .438 & .004 / .256 & .403 / .386 \\
Fixed threshold & .054 / .281 & 1.439 / .538 & .096 / .279 & .716 / .433 & .007 / .258 & .384 / .385 \\
\rcr{} & .063 / .285 & 1.515 / .546 & .099 / .279 & .778 / .442 & .012 / .258 & .394 / .385 \\
Exact every step & .063 / .282 & 1.538 / .553 & .111 / .281 & .784 / .445 & .005 / .257 & .387 / .388 \\
\bottomrule
\end{tabular}
\caption{Reference-based CIDEr / ROUGE-L (budget 64). Dynamic methods are not separated reliably by references.}
\label{tab:ref}
\end{table*}

\section{Qualitative Examples}
\label{app:qual}
Omitted or confabulated details are especially costly in long descriptions, which are already more prone to hallucination than short answers~\citep{zheng2025longer}. Table~\ref{tab:qual} shows two Qwen2-VL DetailCaps images, chosen from the two tails of the per-sample fidelity difference between \rcr{} and InfiniGen (so neither is typical). Outputs are shortened with [\dots]; the shared opening sentence is omitted where identical. Over the 300 test images, \rcr{} is closer to the uncompressed output than InfiniGen by more than 0.1 ROUGE-L on 70 images and farther by more than 0.1 on 41; the mean paired difference is +2.8 points (Table~\ref{tab:main}).

\begin{table*}[t]
\centering
\small
\begin{tabular}{p{0.13\textwidth}p{0.80\textwidth}}
\toprule
\multicolumn{2}{l}{\textbf{Image 2475 (forest with a dog), favourable to \rcr{}}} \\
\midrule
Uncompressed & [\dots] The sunlight appears to be coming from behind the trees, creating a silhouette effect on the left side of the image, \textbf{where a dog is visible. The dog is standing still, facing towards the right side of the image}, and appears to be observing something in the distance. [\dots] \\
BACON (.33) & [\dots] with a dog prominently featured in the foreground. The dog appears to be a medium to large breed, possibly a Golden Retriever [\dots] capturing a moment of quiet companionship [\dots] \\
InfiniGen (.60) & [\dots] creating a silhouette effect on the left side of the image. The trees are mostly coniferous [\dots] The forest floor is not clearly visible [\dots] \emph{(no dog is mentioned)} \\
\rcr{} (1.00) & Identical to the uncompressed output. \\
\midrule
\multicolumn{2}{l}{\textbf{Image 3142 (wristwatch), unfavourable to \rcr{}}} \\
\midrule
Uncompressed & [\dots] The watch face is circular and also silver-colored, with a white background. \textbf{The hour markers on the watch face are small, circular, and also silver-colored} [\dots] The watch hands are thin and silver-colored [\dots] \\
BACON (.45) & [\dots] The watch face is encrusted with numerous small, round, clear gemstones [\dots] the hour markers are likely to be small, possibly diamond-like [\dots] \\
InfiniGen (1.00) & Identical to the uncompressed output. \\
\rcr{} (.51) & [\dots] The watch face itself is circular and features a white background with \textbf{black hour markers and hands} [\dots] a small, round, and possibly decorative element at the top of the case [\dots] \\
\bottomrule
\end{tabular}
\caption{Outputs on two images; numbers are fidelity to the uncompressed output.}
\label{tab:qual}
\end{table*}

\section{Negative Results}
\label{app:neg}
\begin{itemize}\setlength\itemsep{1pt}
  \item Scale-free thresholds relative to the residual drift after the last refresh ($\kappa=2,3$) almost never fire (0.5\% refresh).
  \item Normalising drift by total prefix attention does not separate low-drift samples (rank correlation 0.98 with absolute drift).
  \item Geometric expansion of the first look-ahead position along the reading direction lowers TextVQA accuracy by 1.9--2.4 points; its added tokens have precision 7--20\%, and key-similarity gating fails as well (precision .106--.146).
  \item Adaptive look-ahead lies on the same accuracy--cost curve as fixed look-ahead; look-ahead gains on short answers are largely evaluation leakage (on Qwen2 TextVQA, 58\% of answers are complete within a 2-step look-ahead).
  \item Cross-layer averaging of last-query attention helps Qwen2 VQA slightly (+1.6 / +1.2 on TextVQA / DocVQA) but not LLaVA, and resembles CLAA~\citep{claa2026}.
\end{itemize}

\section{Video Description (Pre-registered)}
\label{app:video}

\paragraph{Registration and setup.} Before any compressed-method video output was produced (only two full-KV smoke outputs had been seen), we registered the data, models, methods, the fixed-threshold headline rule of Appendix~\ref{app:q3}, the primary comparison (fidelity, paired bootstrap against InfiniGen) and that all results are reported. Data: VideoDetailCaption from lmms-eval~\citep{lmmseval2024} (499 ActivityNet~\citep{activitynet2015} videos with human detailed descriptions), 300 videos sampled evenly, benchmark prompt unchanged, 32 frames per video, frame size 448$\times$252 for Qwen2-VL and 512$\times$288 for Qwen3-VL (the same number of visual tokens per frame), \texttt{max\_new\_tokens}=512, greedy decoding. Mean prefixes are 2,205 and 2,342 tokens; uncompressed descriptions have a median of 123 and 303 words. The budget (64), BACON prefill and \rcr{}'s configuration are those of the image experiments; LLaVA-NeXT has no video input. Table~\ref{tab:video} lists all runs.

\begin{table}[h]
\centering
\small
\setlength{\tabcolsep}{2.5pt}
\resizebox{\columnwidth}{!}{%
\begin{tabular}{lcc}
\toprule
VideoDetailCaption & Qwen2-VL & Qwen3-VL \\
\midrule
BACON & .363\up{$-$7.9 [$-$9.2, $-$6.7]} & .266\up{$-$12.6 [$-$13.5, $-$11.7]} \\
Two-stage Quest, page 8 & .423\up{$-$2.0 [$-$3.3, $-$0.7]} & .365\up{$-$2.7 [$-$3.6, $-$1.9]} \\
InfiniGen & .442 & .392 \\
Fixed $\tau{=}.12$ & .423 \pt{7\%}\up{$-$1.9 [$-$3.1, $-$0.7]} & .369 \pt{6\%}\up{$-$2.3 [$-$3.2, $-$1.3]} \\
Fixed $\tau{=}.08$ & \textbf{.438} \pt{18\%}\up{$-$0.5 [$-$1.7, +0.9]} & .394 \pt{12\%}\up{+0.2 [$-$0.7, +1.1]} \\
Fixed $\tau{=}.05$ & .461 \pt{35\%}\up{+1.9 [+0.6, +3.2]} & \textbf{.406} \pt{23\%}\up{+1.4 [+0.6, +2.2]} \\
Fixed $\tau{=}.03$ & .459 \pt{56\%}\up{+1.6 [+0.1, +3.2]} & .417 \pt{39\%}\up{+2.5 [+1.7, +3.4]} \\
\rcr{} ($\rho{=}.4$) & .452 \pt{40\%}\up{+0.9 [$-$0.6, +2.4]} & .412 \pt{40\%}\up{+2.0 [+1.1, +2.8]} \\
Exact every step & .466\up{+2.4 [+0.9, +3.8]} & .420\up{+2.8 [+1.9, +3.7]} \\
\bottomrule
\end{tabular}}
\caption{Video fidelity (300 videos, budget 64). Parentheses: refresh rate; small numbers: paired difference to InfiniGen with 95\% CI. Bold: fixed-threshold headline chosen by the registered rule.}
\label{tab:video}
\end{table}

\paragraph{\rcr{} versus the fixed threshold.} At the nearest refresh rates, \rcr{} minus the fixed threshold is $-$1.0 [$-$2.4, +0.5] ($\tau{=}.05$, 35\%) and $-$0.7 [$-$2.2, +0.8] ($\tau{=}.03$, 56\%) on Qwen2-VL, and $-$0.5 [$-$1.4, +0.3] ($\tau{=}.03$, 39\%) and +0.5 [$-$0.3, +1.4] ($\tau{=}.05$, 23\%) on Qwen3-VL. By the registered rule, the fixed-threshold headline is $\tau{=}.08$ on Qwen2-VL and $\tau{=}.05$ on Qwen3-VL, which \rcr{} exceeds by 1.4 and 0.6 points at about twice their refresh rate. The corresponding image comparisons at similar rates are +1.1 [$-$0.0, +2.3] (LLaVA DetailCaps, fixed 35\%), +1.3 [$-$0.3, +2.9] (Qwen3 DetailCaps, 34\%), +3.6 [+1.9, +5.2] (Qwen2 TextCaps, 29\%), +2.9 [+1.2, +4.6] (LLaVA TextCaps, 28\%) and +0.5 [$-$0.8, +1.7] (Qwen3 TextCaps, 44\%); note that \rcr{} also differs in detector and candidate re-selection.

\paragraph{Drift heterogeneity.} Table~\ref{tab:spread} summarises how much per-sample mean drift varies under the fixed threshold at its headline setting.

\begin{table}[h]
\centering
\small
\setlength{\tabcolsep}{3pt}
\begin{tabular}{lccc}
\toprule
Setting & CV & p90/p10 & refresh $<$5\% \\
\midrule
Qwen2-VL DetailCaps & 0.71 & 7.0 & 21\% \\
Qwen3-VL DetailCaps & 0.78 & 8.0 & 45\% \\
LLaVA-NeXT DetailCaps & 0.16 & 1.5 & 0\% \\
Qwen2-VL video & 0.18 & 1.5 & 1\% \\
Qwen3-VL video & 0.14 & 1.4 & 0\% \\
\bottomrule
\end{tabular}
\caption{Spread of per-sample mean drift: coefficient of variation, 90th/10th percentile ratio, and share of samples refreshing on fewer than 5\% of layer-steps.}
\label{tab:spread}
\end{table}

\section{Object-Level Evaluation}
\label{app:capture}
We parse each output with the factual scene-graph parser used by CAPTURE~\citep{detailcaps2024} (FLAN-T5-base) and match objects, attributes and relations with CAPTURE's exact, synonym and embedding matching. We report CAPTURE against the three human references of each DetailCaps image, object precision and recall against the references (averaged over references), and object precision, recall and F1 against the uncompressed output of the same model. Against the uncompressed output, lower precision indicates added objects and lower recall indicates omitted ones. Table~\ref{tab:capture} lists all methods on the 300 test images; the fixed threshold is the headline setting of Table~\ref{tab:main}.

\begin{table*}[t]
\centering
\small
\setlength{\tabcolsep}{3.2pt}
\resizebox{\textwidth}{!}{%
\begin{tabular}{lccccccccc}
\toprule
& \multicolumn{3}{c}{Qwen2-VL-7B} & \multicolumn{3}{c}{LLaVA-NeXT-7B} & \multicolumn{3}{c}{Qwen3-VL-8B} \\
\cmidrule(lr){2-4}\cmidrule(lr){5-7}\cmidrule(lr){8-10}
Method & CAPTURE & P/R ref & P/R/F1 full & CAPTURE & P/R ref & P/R/F1 full & CAPTURE & P/R ref & P/R/F1 full \\
\midrule
BACON & .602 & .659/.634 & .733/.702/.708 & .554 & .652/.539 & .698/.632/.651 & .611 & .602/.720 & .779/.721/.744 \\
Two-stage Quest & .631 & .684/.676 & .802/.790/.786 & .593 & .694/.586 & .774/.736/.745 & .624 & .594/.770 & .818/.813/.812 \\
InfiniGen & .631 & .680/.676 & .823/.815/.811 & .590 & .716/.573 & .785/.708/.733 & .627 & .596/.778 & .821/.819/.816 \\
Fixed threshold & .634 & .682/.686 & .809/.811/.802 & .592 & .693/.594 & .772/.737/.745 & .626 & .592/.777 & .808/.809/.805 \\
\rcr{} & .635 & .687/.682 & .835/.827/.824 & .598 & .694/.597 & .778/.756/.758 & .626 & .594/.777 & .825/.827/.823 \\
Exact every step & .634 & .684/.681 & .841/.835/.832 & .597 & .700/.596 & .783/.754/.759 & .626 & .593/.776 & .827/.831/.826 \\
Uncompressed & .636 & .683/.684 & 1/1/1 & .607 & .700/.618 & .997/.997/.997 & .628 & .594/.780 & 1/1/1 \\
\bottomrule
\end{tabular}}
\caption{Object-level evaluation on DetailCaps (300 images, budget 64). ``ref'': against the human references; ``full'': against the uncompressed output. LLaVA's uncompressed row is below 1 because one of its outputs yields no parsed objects and scores 0.}
\label{tab:capture}
\end{table*}

Paired differences of \rcr{} minus InfiniGen (points, 95\% CI): object F1 against the uncompressed output +1.3 [+0.2, +2.3] (Qwen2-VL), +2.5 [+1.4, +3.5] (LLaVA-NeXT), +0.7 [$-$0.1, +1.5] (Qwen3-VL); recall +1.2 [$-$0.2, +2.6], +4.8 [+3.3, +6.2], +0.9 [$-$0.0, +1.8]; precision +1.2 [$-$0.1, +2.5], $-$0.7 [$-$2.1, +0.7], +0.3 [$-$0.6, +1.3]; CAPTURE +0.4 [+0.1, +0.7], +0.8 [+0.4, +1.3], $-$0.1 [$-$0.3, +0.2]; object precision against the references +0.7 [$-$0.0, +1.4], $-$2.2 [$-$3.2, $-$1.2], $-$0.2 [$-$0.7, +0.4]. Against the fixed threshold, object F1 differs by +2.2 [+1.0, +3.3], +1.3 [+0.3, +2.3] and +1.9 [+1.1, +2.6].

\section{Implementation Details}
\label{app:impl}
All methods are implemented in a single module patched into the attention forward of each model (Hugging Face Transformers; Qwen2-VL and LLaVA in fp16, Qwen3-VL in bf16). \rcr{}'s configuration is: budget 64 ($c=W=32$), BACON prefill, $\rho=0.4$, $\tau_0=0.03$, $\eta=0.05$, $R=8$ detector channels, $4c$ candidates, and no EMA (decay 0; Appendix~\ref{app:fixed}). The threshold is reset for every sample, so results do not depend on sample order. Summing Eq.~\eqref{eq:rate} over the $N$ layer-level checks of a sample gives $\frac{1}{N}\sum_i m_i-\rho=(\log\tau_N-\log\tau_0)/(\eta N)$, so the realised rate approaches $\rho$ only as $N$ grows, consistent with the overshoot on short TextCaps outputs. DetailCaps uses 300 images sampled evenly from DetailCaps-4870 with \texttt{max\_new\_tokens}=512; TextCaps uses the first 500 validation images with 64. Mean prefix lengths are 852 / 1012 (Qwen2 DetailCaps / TextCaps), 1679 / 2310 (LLaVA) and 639 / 764 (Qwen3-VL). With refresh disabled, every model reproduces BACON token-for-token. With a mask-conditioned read from host memory, \rcr{}'s 64 vectors per layer, KV head and step amount to about 1.8\,MB per token for Qwen2-VL-7B in fp16, or 75\,$\mu$s at 25\,GB/s. Code, logs and the registration files will be released.

\end{document}